\documentclass[letterpaper, 10pt, conference]{ieeeconf}
\usepackage{graphicx} 
\usepackage{algorithm}

\IEEEoverridecommandlockouts

\usepackage{amsmath}
\usepackage{amssymb}
\usepackage{amsthm}
\usepackage{algorithm}
\usepackage{algpseudocode}
\usepackage{etoolbox} 
\usepackage[hidelinks]{hyperref} 
\usepackage[dvipsnames]{xcolor}
\usepackage{optidef}    
\usepackage{cite}

\newcommand{\arxiv}[2]{\ifbool{\usingarxiv}{#1}{#2}} 

\newcommand{\norm}[1]{\rVert #1\lVert}                  

{                                                       
    \theoremstyle{plain}

    \theoremstyle{remark}

    \theoremstyle{definition}

}

\newcommand{\bbR}{\mathbb{R}}

\newcommand{\calX}{\mathcal{X}}
\newcommand{\calU}{\mathcal{U}}

\newcommand{\set}[2]{\{#1 \: | \: #2\}}

\title{\LARGE \bf Locomotion on Constrained Footholds via \\ Layered Architectures and Model Predictive Control}
\author{
    Zachary Olkin$^{1}$ and Aaron D. Ames$^{1}$%
    \thanks{This research is supported by Technology Innovation Institute (TII) and the National Science Foundation Graduate Research Fellowship.}
    \thanks{$^{1}$Authors are with the Department of Control and Dynamical Systems, California Institute of Technology, Pasadena CA 91125, U.S.A. \texttt{\{zolkin, ames\}@caltech.edu}.}
    \thanks{The supplementary video can be found here: \url{https://youtu.be/pTIWqsFODQk}}
}

\begin{document}
 
\maketitle

\begin{abstract}    
    Computing stabilizing and optimal control actions for legged locomotion in real time is difficult due to the nonlinear, hybrid, and high dimensional nature of these robots. The hybrid nature of the system introduces a combination of discrete and continuous variables which causes issues for numerical optimal control. To address these challenges, we propose a layered architecture that separates the choice of discrete variables and a smooth Model Predictive Controller (MPC). The layered formulation allows for online flexibility and optimality without sacrificing real-time performance through a combination of gradient-free and gradient-based methods. The architecture leverages a sampling-based method for determining discrete variables, and a classical smooth MPC formulation using these fixed discrete variables. We demonstrate the results on a quadrupedal robot stepping over gaps and onto terrain with varying heights. In simulation, we demonstrate the controller on a humanoid robot for gap traversal. The layered approach is shown to be more optimal and reliable than common heuristic-based approaches and faster to compute than pure sampling methods.
\end{abstract}

\section{Introduction}
Legged robots make and break contact with the environment to move around and interact with the world. Humanoid robots also have arms with the hope that they will be used to manipulate objects and assist in challenging locomotion scenarios. The choice of contact with the environment is what gives these robots the potential to navigate through complex terrain and manipulate objects. Yet, choosing contacts is a major challenge in robotics due to its inherently hybrid nature. Locomotion over complex terrain is of particular interest for legged robots because this is where they excel over many other designs.

Dynamic locomotion over terrain, like ``stepping stones", is difficult for a number of reasons. These system are nonlinear, high-dimensional, and hybrid; all of these together mean that solving the full nonlinear hybrid optimization problem is prohibitively slow for real-time use. On the other hand, recently, Model Predictive Control (MPC) has generated highly dynamic and complex motions for both quadrupeds and humanoids by fixing contact sequences \cite{li_cafe-mpc_2024, khazoom_tailoring_2024, sleiman_unified_2021, zhou_momentum-aware_2022, galliker_planar_2022}. In this paper, we propose a method that preserves the computation speed and dynamic capability of MPC while also optimizing over discrete choices like stepping stones. Our method exploits a layered architecture \cite{olkin_layered_2025} where the discrete choices are chosen in one layer and fixed for the real-time MPC.

\begin{figure}[!t]
    \centering
    \includegraphics[width=0.95\linewidth]{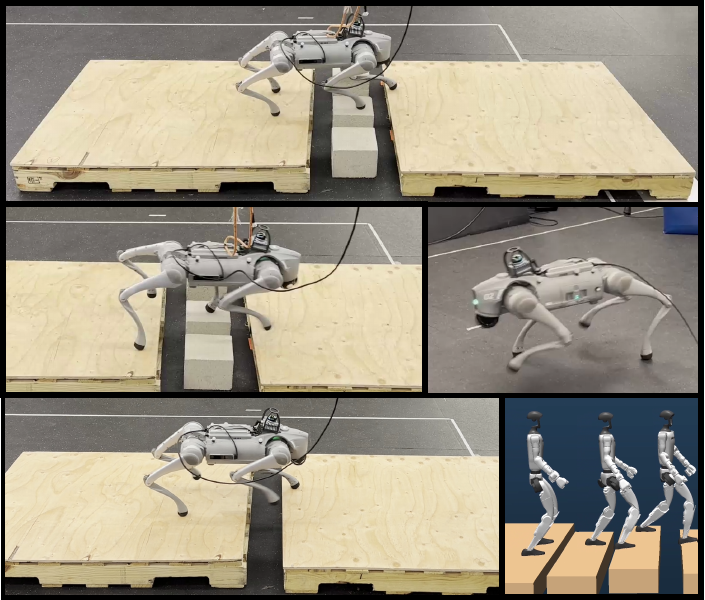}
    \caption{Demonstration of the layered controller dynamically navigating over complex terrain and locomoting effectively on flat ground. The controller is both versatile in the terrain and actions it can stabilize and fast enough for real-time control.}
    \label{fig:hero_fig}
    \vspace{-5mm}
\end{figure}

\subsection{Related Works}
\begin{figure*}
    \vspace{3mm}
    \centering
    \includegraphics[width=1\linewidth]{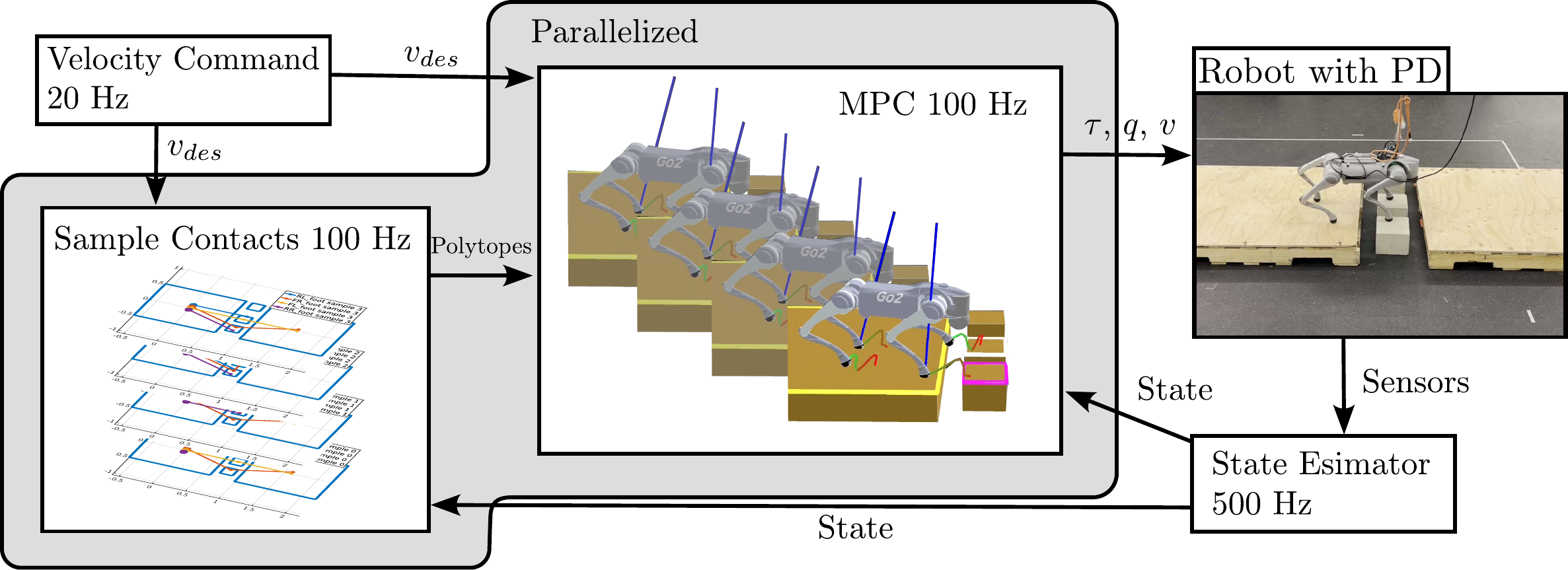}
    \caption{Diagram of the control architecture. Velocity commands from the user are used for sampling and go into the MPC cost. A set of contacts (terrain surfaces) are sampled and passed into the MPC as polytope constraints. The MPC computes full order trajectories in parallel for each set of polytopes, selecting the lowest cost option to use. The MPC then outputs feedforward torques, $\tau$, joint configurations, $q$, and joint velocities $v$ as shown in the middle box. In the MPC box, the feet trajectories are shown in a green-red gradient where green is closer in time. The polytopes to be stepped on are shown highlighted in yellow, white, and pink (color coded to each foot). The blue lines show the force computed by the MPC. A state estimator outputs the state at 500 Hz.}
    \label{fig:control_arch}
    \vspace{-2mm}
\end{figure*}
Mixed-integer optimization has been used to solve the hybrid optimization problem. Authors of \cite{aceituno-cabezas_simultaneous_2017} plan gait transitions, motions, and terrain selection at once using mixed-integer convex constraints. Alternatively, \cite{deits_footstep_2014} plans the footstep locations and orientations by solving a mixed-integer convex program which can then be used for the motion generation. These approaches are computationally intensive, and thus cannot be used for real-time control.

Recently, a number of works have used MPC, accompanied by a heuristic, for locomotion through segmented terrain \cite{grandia_multi-layered_2021, grandia_perceptive_2022}. The heuristic selects the terrain to step on and passes this as a constraint to the MPC. These algorithms are computationally efficient, but require a good heuristic. If there is no quality heuristic, say for using a humanoid's arms, then these concepts may not perform well.

Contact Implicit MPC (or CIMPC), attempts to solve for contacts and motion in a single, gradient based optimization. There have been a number of approaches to this problem using a variety of approximations and optimization techniques. Some works have used smooth, non-physical, ground reaction forces \cite{kurtz_inverse_2023, neunert_whole-body_2018}. On the other hand, a number of methods use hard contacts  \cite{kong_hybrid_2022, kong_ilqr_2021, kim_contact-implicit_2023, posa_direct_2013}. Alternatively, a gradient based method with optimization based contact models was proposed in \cite{cleach_fast_2023}. In general, these solvers are subject to poor numerical conditioning \cite{wensing_optimization-based_2022} and slower solve speeds than fixed contact MPC.

Recently, there has been a push to use pure sampling based, gradient free, optimization for full order control of legged systems \cite{howell_predictive_2022, xue_full-order_2024, alvarez-padilla_real-time_2024, pan_model-based_2024}. A common optimization scheme is Model Predictive Path Integral (MPPI) control \cite{williams_model_2017} which uses random sampling of control actions and simulation rollouts to optimize inputs. These methods are particularly enticing due to the ability to automatically discover contact sequences and interact with the environment. To the best of our knowledge, no full-order sampling based method has been shown for real-time control of a humanoid robot.

\subsection{Contributions}
In this work, we propose a layered architecture with MPC to locomote through complex terrain. The layered architecture has two components: (1) a sampling-based, gradient-free method for selecting terrain for locomotion and, (2) a full order nonlinear MPC controller outputting PD setpoints and feed forward torques. The layered control architecture allows the robot to select which footholds to use in real time while considering the closed loop performance of the system for a variety of terrain options. In addition, the MPC is formulated to optimize on the quaternion tangent space rather than using Euler angles as is commonly done in MPC.

The controller is faster than pure sampling-based and CIMPC methods, and it can be run in real time for a humanoid. The proposed controller also outperforms the Raibert heuristic \cite{raibert_legged_2000} for selecting terrain when traveling at high speeds, as shown in simulation experiments. Further, the controller does not require heuristics, which allows it to generalize to other tasks beyond stepping stones. We demonstrate the performance of the controller on the Unitree Go2 quadruped robot crossing over a variety of terrains and locomoting on flat ground, as shown in Fig. \ref{fig:hero_fig}. Simulation results shown the efficacy of the approach on the Unitree G1 humanoid crossing over complex terrain.

\section{Layered Control and MPC Formulation}
The architecture has two major parts: a gradient free algorithm that chooses the discrete variables (i.e. selects which pieces of terrain to use), and a smooth MPC that fixes these discrete variables and is formulated as a classical smooth Nonlinear Program (NLP). We choose the discrete variables through a sampling based method: given the velocity command to the robot, we propagate forward a nominal stance foot location and sample the terrain around that location. Then, the fixed mode MPC is solved in parallel for every sample. The lowest cost MPC solve is used as the controller output. To enforce the terrain sample in the fixed mode MPC, we use a polytope constraint on the end effector during stance, similar to \cite{grandia_perceptive_2022}. The entire architecture is shown in Fig. \ref{fig:control_arch}.

\subsection{Fixed-Mode MPC Formulation}
Once we have fixed the discrete variables, like terrain, for a given MPC solve, we can write the fixed mode problem as follows:
\begin{mini!}[0]
{x_{[0:N]}, \; u_{[0:N-1]}}{J(x, u)}
{\label{eq:nlp_mpc}}{}
\addConstraint{x_0}{= \hat{x} \label{eq:ic_eq_fixed_mode_mpc}}
\addConstraint{x_{k+1}}{= g(x_k, u_k) \;}{k = 0...N-1}
\addConstraint{x_k}{\in \calX_k}{k = 0...N}
\addConstraint{u_k}{\in \calU} {k = 0...N-1}
\end{mini!}
where $\hat{x}$ denotes the current state of the system and $N$ denotes the number of nodes in the MPC time horizon. The subscripts represent indexing into the trajectory at a given node. The discrete dynamics of the system are given by $g(x_k, u_k)$. The state and input constraints are given by $\calX$ and $\calU$ and are described in more detail below.

As written in \eqref{eq:nlp_mpc}, the optimization problem is a large NLP. To facilitate real-time control, the MPC formulation uses the idea of Real Time Iterations \cite{diehl_real-time_2005} where only a single quadratic program (QP) is solved at each step. This QP can be viewed as a single iteration of a Sequential Quadratic Programming (SQP) type method. The Linear-Quadratic (LQ) approximation used in the QP is generated by linearizing about the current state and input trajectory, denoted respectively as $\bar{x}$ and $\bar{u}$.

To embed the contact schedule into the NLP, at each node we associate a set of binary variables $\gamma^l_k \in \{0,1\}$ that denotes if end effector $l$ at node $k$ is in or out of contact (a 1 indicating in contact). For the results of this paper we use a fixed contact schedule so that $\gamma^l_k$ is not an optimization variable. By using this parameterization, the structure of the MPC never changes from solve to solve \cite{khazoom_tailoring_2024}. 

For each QP, the decision variables are not the state and input for the resulting trajectory, but rather they are the change in the states and inputs to get to the next trajectory from the current trajectory \cite{jackson_planning_2021}. By making the decision variables relative to the current trajectory, we can optimize the orientation of the robot on the tangent space of the orientation manifold and avoid singularities associated with Euler angles. Therefore, the state decision variables take the following form at each node $k$: $x_k = [\delta q_k, \delta v_k] \in \bbR^{n_x - 1}$ where $n_x$ denotes the size of the state variables at each node. Note that the dimension of $x_k$ is $n_x - 1$ and not $n_x$ because $\delta q_k$ does not have a quaternion element and instead lies on the tangent space of the manifold which has lower dimension. The input variables take the form $u_k = [\delta \tau_k, \delta F_k] \in \bbR^{n_u}$ with $n_u$ the size of the inputs, $\tau$ the torques at the joints, and $F_k$ the forces on the end effectors.  The trajectory used for linearization takes the form $\bar{x}_k = [\bar{q}_k, \bar{v}_k]^T \in \bbR^{n_x}$ and $\bar{u}_k = [\bar{\tau}_k, \bar{F}_k]^T \in \bbR^{n_u}$.

To solve the QP, we use the HPIPM solver \cite{frison_hpipm_2020} which requires a optimal control block structure. This will effect how the constraints are formulated in the following sections.

\subsubsection{Dynamics Constraints}
We use inverse dynamics for the dynamics constraints. Denote the inverse dynamics as
\begin{equation}
    \tau_k = f(q_k, v_k, a_k, F_k).
\end{equation}

We can approximate $a$ by finite differencing: $a = \frac{v_{k+1} - v_k}{\delta t_k}$. To fit into the optimal control structure necessitated by HPIPM, we need to write the discretized dynamics explicitly, i.e., with $\delta v_{k+1}$ and $\delta q_{k+1}$ as the result of a function of $u_k$ and $x_k$. First, note that the inverse dynamics can be re-written in terms of $x_k$ and $u_k$ as follows:
\begin{equation}
    \bar{\tau}_k + \delta \tau_k = f(\bar{q}_k \oplus \delta q_k, \bar{v}_k + \delta v_k, \bar{a}_k + \delta a_k, \bar{F}_k + \delta F_k).
    \label{eq:full_id}
\end{equation}
We define $\oplus$ as the right quaternion addition for the elements of the vector corresponding to a quaternion and normal linear addition for all the other elements. The quaternion addition is just $\mathbf{q} \circ \text{Exp}(\delta \mathbf{q})$ with $\mathbf{q}$ a quaternion and $\delta \mathbf{q}$ a vector element and $\circ$ is quaternion multiplication. See \cite{sola_micro_2021} for more details.

Now, we re-write the right hand side of \eqref{eq:full_id} as a function of just the differential terms and substitute the finite difference acceleration approximation: 
\begin{multline*}
    \bar{f}(\delta q_k, \delta v_k, \delta v_{k+1}, \delta F_k) := f(\bar{q}_k \oplus \delta q_k, \bar{v}_k + \delta v_k, \\ \bar{a}_k + \delta a_k, \bar{F}_k + \delta F_k).
\end{multline*}

Then, dropping the function arguments for the sake of presentation, and linearizing the inverse dynamics yields
\begin{multline*}
    \bar{\tau}_k + \delta \tau_k \approx \nabla_{\delta q_k} \bar{f} \delta q_k + \nabla_{\delta v_k} \bar{f} \delta v_k + \\ \nabla_{\delta F_k} \bar{f} \delta F_k + \nabla_{\delta v_{k+1}} \bar{f} \delta v_{k+1} + \bar{f}.
\end{multline*}
Inverting the Jacobian in front of $\delta v_{k+1}$ and re-arranging gives the desired structure:
\begin{multline}
    \delta v_{k+1} \approx -\nabla_{\delta v_{k+1}} \bar{f}^{-1}(\nabla_{\delta q_k} \bar{f} \delta q_k + \nabla_{\delta v_k} \bar{f} \delta v_k + \\ \nabla_{\delta F_k} \bar{f} \delta F_k - \bar{\tau}_k - \delta \tau_k + \bar{f}).
\end{multline}

We still need the integration constraint relating the configuration to the velocity. The nonlinear update is given as $q_{k+1} = q_k \oplus v_k \delta t$. This can be written as a function of the differentials: $\bar{f}_I(\delta q_k, \delta v_k) :=  \bar{q}_{k} \oplus \delta q_k \oplus (\bar{v}_k + \delta v_k)\delta t$. This gives us
\begin{equation*}
    \bar{q}_{k+1} \oplus \delta q_{k+1} = \bar{f}_I(\delta q_k, \delta v_k).
\end{equation*}
Then, with suppressed function arguments, linearizing yields
\begin{multline*}
    \nabla_{\delta q_{k+1}} (\bar{q}_{k+1} \oplus \delta q_{k+1}) \delta q_{k+1} + \bar{q}_{k+1} \approx \nabla_{\delta q_k}\bar{f}_I \delta q_k + \\ \nabla_{\delta v_k}\bar{f}_I \delta v_k + \bar{f}_I,
\end{multline*}
which can then be re-arranged
\begin{multline}
        \delta q_{k+1} \approx \nabla_{\delta q_{k+1}} (\bar{q}_{k+1} \oplus \delta q_{k+1})^{-1} (-\bar{q}_{k+1} + \\ \nabla_{\delta q_k}\bar{f}_I \delta q_k + \nabla_{\delta v_k}\bar{f}_I \delta v_k + \bar{f}_I).
        \label{eq:linearize_integration}
\end{multline}
Eq. \eqref{eq:linearize_integration} gives the linearized version of the integration dynamics constraint in the optimal control structure necessitated by HPIPM.
\subsubsection{Inequality Constraints}
We apply box constraints for the configuration, velocity, torque, and ground reaction forces:
\begin{align*}
    q_{lb} &\leq \bar{q}^j_k + \delta q^j_k \leq q_{ub} \\
    v_{lb} &\leq \bar{v}^j_k + \delta v^j_k \leq v_{ub} \\
    \tau_{lb} &\leq \bar{\tau}_k + \delta \tau_k \leq \tau_{ub} \\
    \gamma_k^l F_{lb} &\leq \bar{F}_k + \delta F \leq \gamma_k^l F_{ub}.
\end{align*}
The $q^j$ and $v^j$ indicate the actuated joints and and not the floating base. The ground reaction forces are multiplied by the contact binary to set the force to zero in the swing phase. A nonzero minimum ground reaction force is used to help prevent foot slippage during contact.

A linearized friction cone constraint is used for end effectors in contact:
\begin{align*}
    |\gamma_k^l(\bar{F}_k + \delta F_k)^x| &\leq \mu \gamma_k^l(\bar{F}_k + \delta F_k)^z \\
    |\gamma_k^l(\bar{F}_k + \delta F_k)^y| &\leq \mu \gamma_k^l(\bar{F}_k + \delta F_k)^z
\end{align*}
where the $x$, $y$, and $z$ superscripts denote the portions of the force in those directions and $\mu$ is the friction coefficient.

Collision constraints are enforced as configuration-only constraints. Pairs of spheres of a given radius are drawn around two locations on the robot and the following constraint is used: $0 \leq R_1 + R_2 - \norm{d_1(q_k) - d_2(q_k)}$ where $R_1$ and $R_2$ are the radius of the spheres, $d_1(q_k)$ and $d_2(q_k)$ are functions of the configuration returning 3-vectors representing the global position of the two locations. In the same vein as the dynamics constraints, first these must be written in terms of the differential, then they are linearized. For the rest of the constraints we will only write the nonlinear version with the understanding that they are linearized to be used in the MPC.

Foot placement constraints are used to ensure that the robot avoids gaps, and steps on the desired terrain. We model steppable regions as polytopes: $\set{y}{Ay \leq b}$. Then, we enforce that the end effector position is inside this region. Therefore the constraint becomes
\begin{equation}
    \gamma_k^l A^l_k r^l(\bar{q}_k \oplus \delta q_k) \leq \gamma_k^l b^l_k.
\end{equation}
where we use the notation $r^l(q)$ to denote the global 3-D position (determined via forward kinematics) of end effector $l$. $A$ and $b$ can be different at each node for each end effector and thus have the corresponding sub and superscripts.

\subsubsection{Equality Constraints}
An equality constraint on the z-height of each foot was imposed in the MPC. This constraint ensures that the foot does not make contact with the ground during the swing phase. The constraint takes the form $r^{l,z}(\bar{q}_k \oplus \delta q_k) = z_k$ where $z_k$ gives the desired swing height at node $k$.

A no-slip constraint is added to each end effector:
\begin{equation}
    \gamma_k^l r^l(\bar{q}_k \oplus \delta q_k, \bar{v}_k + \delta v_k) = 0.
\end{equation}
This constraint ensures that when a given end effector is in contact that there is no velocity at that point. Here, with a slight abuse of notation, we let $r^l$ denote the first order forward kinematics, i.e. the velocity of end effector $l$ in the the local frame.

\subsubsection{Cost Function}
The full nonlinear cost function is composed of a few terms: state tracking/regularization, input regularization, end effector tracking, and a terminal cost. The cost is:
\begin{multline}    
    J_{nl}(x,u) = \sum_{k = 0}^{N-1} \big( \norm{\hat{x}_k \ominus x_k}^2_{W_x} + \norm{\hat{u}_k - u_k}^2_{W_u} + \\ 
    \sum_{l = 0}^{N_l} \big(\norm{\hat{r}^l - r^l(q_k)}^2_{W_r} \big)\big) + \eta \norm{\hat{x}_N \ominus x_N}^2_{W_{x_N}}
    \label{eq:cost_function}
\end{multline}
where the hats ($\hat{\cdot}$) represent desired quantities, $W_{(\cdot)}$ represent weights, and $\eta$ is the terminal cost weight. This cost function is not quadratic for two reasons: (1) the subtraction $\hat{x}_k \ominus x_k$ includes a quaternion subtraction that is nonlinear, and (2) the function $r^l(q_k)$ is nonlinear. We will use a Gauss-Newton Hessian approximation as this generally works well in practice and exploits the least-squares form of the cost \cite{houska_auto-generated_2011, verschueren_acados_2020}. First, we re-write the state tracking in terms of the differential: $\norm{\hat{x}_k \ominus x_k}^2_{W_x} = \norm{\hat{x}_k \ominus \bar{x}_k \oplus \delta x_k}^2_{W_x}$. Then we can get the Hessian approximation:
\begin{align*}
    H_x &\approx \nabla_{\delta x_k}(\bar{x}_k \oplus \delta x_k)^T W_x \nabla_{\delta x_k}(\bar{x}_k \oplus \delta x_k),
\end{align*}
which is also used on the terminal cost. We can use the same trick on the end effector term:
\begin{align*}
    H_r &\approx \nabla_{\delta q_k}(r^l(\bar{q}_k \oplus \delta q_k))^T W_r \nabla_{\delta q_k}(r^l(\bar{q}_k \oplus \delta q_k)).
\end{align*}

Then we define $J(x,u) \approx J_{nl}(x, u)$ using the Hessian approximations, the true Hessians for the quadratic terms, and the associated linear terms. 

\subsection{Sampling for Gradient Free Decision Making}
To make this fixed-mode MPC, we need a way to choose the discrete variables that do not have a gradient associated with them and therefore cannot be optimized using classical smooth NLP tools. For example, choosing which piece of terrain to step on (i.e. choosing the polytope to use in the polytope constraints), or choosing the contact schedule (i.e. choosing when $\gamma$ is 0 or 1) are both decisions that cannot be optimized in the QP. These types of decisions are very common in legged systems and considering them independently of the dynamics may lead to conservativeness and/or system failure.

To select terrain for locomotion, we use a sampling based method where we evaluate the cost of stepping on each piece of terrain throughout the MPC horizon and choose the lowest cost set of terrain. To do this, we parallelize the MPC computation and choose different polytopes for each MPC solve. Note that this same idea can be applied to the optimizing the contact schedule where $\gamma$ is decided at each node, but in this work we focus on the sampling of terrain. The specifics of the sampling algorithm presented below are thus tailored to that application.

The sampling algorithm works as follows. First, given a desired velocity, we propagate the torso of the robot forward in the $x$-$y$ plane throughout the MPC horizon. Given the resulting torso placement, we check a nominal foot location, i.e., right below the hip at the middle of the stance phase \cite{raibert_legged_2000}. If the nominal foot location is not over a piece of terrain, then the sampling process continues; otherwise, we use the terrain at the nominal location\footnote{Technically, you could choose to sample here even if it is over a piece of terrain (and thus not use the nominal position heuristic), but to reduce the sample space, we choose to take the nominal terrain.}.

When multiple legs have the same contact phases, then we choose to consider their sampling together to prevent re-sampling the same terrain combinations. A ``sample tree" is used to keep track of the possibilities. This tree holds all possible combinations of samples for the given set of feet for a single contact phase. The nodes of the tree represent discrete pieces of terrain. To understand the tree structure better, consider an example where there are two legs with the same contact phases. In this case, the tree will have a depth of three (number of feet plus one): a dummy root node with a number of children equal to the number of terrain options for the first foot, then each of those nodes has a number of children for all the terrain options for the second foot. Therefore, a given branch of the tree corresponds to a valid, unique, terrain selection.

To create the tree, and prevent sampling terrain that is outside of the kinematic limits of the robot, we create a circle in the $x$-$y$ plane centered on the nominal foot location that approximates the kinematic limits of the robot\footnote{Note that any method that gives a true or approximate representation of the kinematic limits can be used here. We find it easy and effective to use a circle about the nominal position.}. Then, we calculate the area of the intersection of each piece of terrain within the circle. Any piece of terrain with non-zero area intersection is added as a node.

Branches are sampled randomly but weighted by the area of the terrain within the kinematic limit to incentivize sampling terrain with large area. A branch is sampled by selecting a node from each layer of the tree (root node excluded). To select a node we sample a number $a \in [0,1]$ uniformly. For all the nodes in the given layer of the tree, we normalize their areas to sum to 1. Then each node is associated with a segment of $[0,1]$ based on its area within the circle. The segment in which $a$ lies is the chosen node, and the process is repeated for the children of that node, etc.... The result of this process is a given branch of the tree. We extract the corresponding polytopes from this branch and store it to be used with MPC. Then that branch is removed so it cannot be sampled again, and the process is repeated for each parallel MPC. Alg. \ref{alg:terrain_sampling} gives an overview of this algorithm.

\begin{algorithm}
\caption{Terrain Sampling}\label{alg:terrain_sampling}
\begin{algorithmic}
\Require number of MPC samples $M$, number of contact phases for each MPC $L$, nominal feet positions to be sampled around, $\zeta_i \in \bbR^{2K} \quad \forall i \leq L$, with $K$ the number of feet in that phase
\While {$i < L$} \Comment{Iterate through each contact phase}
    \State tree = \Call{CreateSampleTree}{$\zeta_i$}
    \While{$j < M$} \Comment{Iterate through each MPC sample}
        \If{tree.\Call{HasChildren}{}}
            \State branch = tree.\Call{SampleBranch}{}
            \State polytope\_bag.\Call{add}{branch.polytopes}
            \State tree.\Call{RemoveBranch}{branch}
        \Else \Comment{More samples than unique terrain options}
            \State  polytope\_bag.add(Randomly select terrain)
        \EndIf
        \State $j = j + 1$
    \EndWhile
    \State $i = i + 1$
\EndWhile
\end{algorithmic}
\end{algorithm}

\begin{figure}
    \vspace{3mm}    
    \centering
    \includegraphics[width=0.95\linewidth]{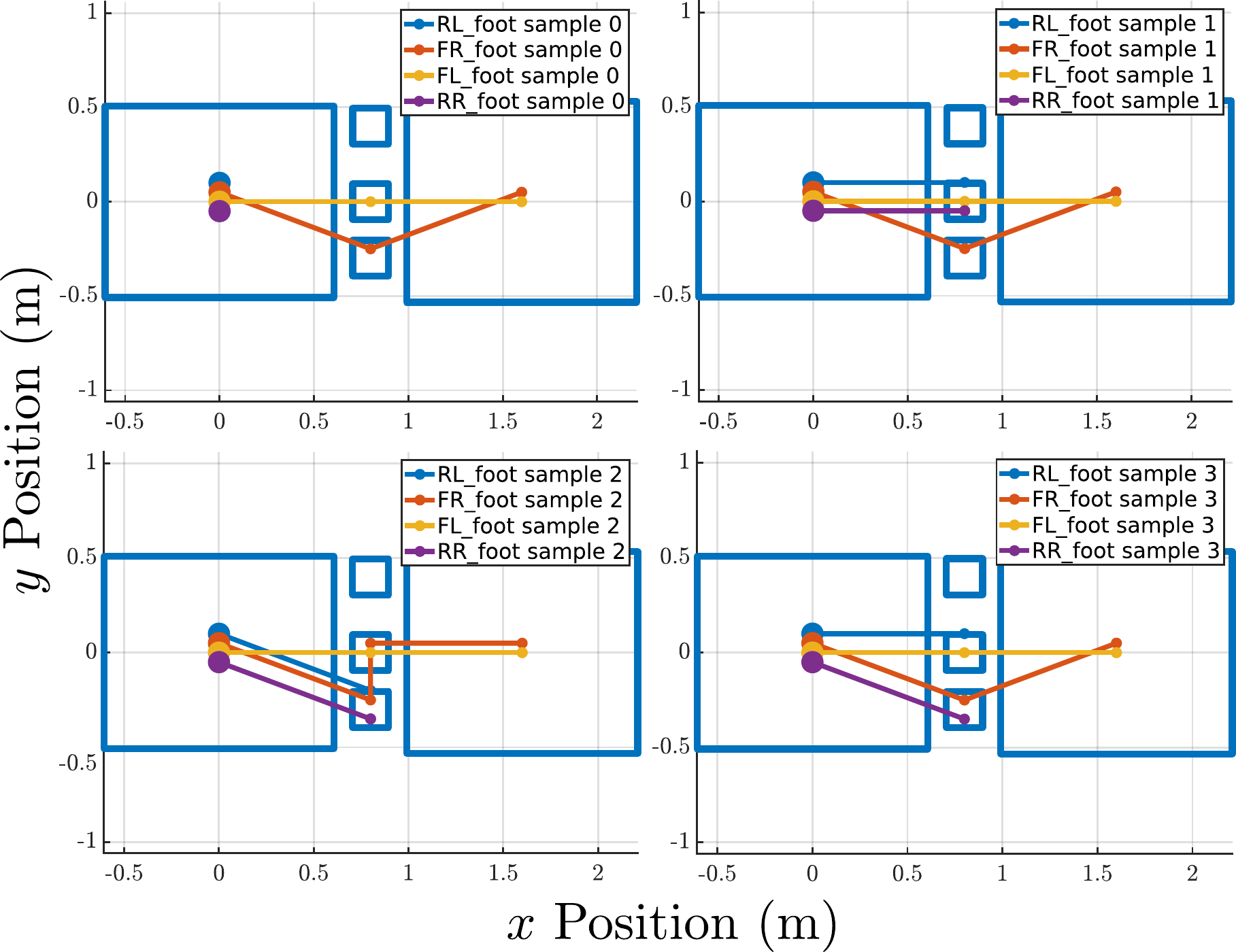}
    \caption{Example result from the sampling. At sampling time, all feet are on the left-most tile (signified by the large circles), and we are plotting all 4 sampled paths. The plotted lines show which blocks are sampled. The plotted position of the circles do not correspond with the feet location within the block.}
    \label{fig:sample_choices_example}
\end{figure}

An example set of samples is shown in Fig. \ref{fig:sample_choices_example}. Each plot shows a different sample to be used with a different MPC solve. We can see that using this sampling approach we are able to explore a variety of terrain options with only four samples.

\section{Implementation}
Implementation of the MPC was done in C++ using HPIPM \cite{frison_hpipm_2020} for the QP solver, Pinocchio \cite{carpentier_pinocchio_2019} for the rigid body dynamics, and CppADCodegen for the derivatives. The fixed-mode MPC is single threaded so that it can be parallelized for the sampling MPC. OpenMP was used to parallelize the fixed-mode MPC.

The MPC is split up into two different phases: the preparation phase and the feedback phase \cite{diehl_real-time_2002}. This is done because some computations can happen before we measure the state (in the preparation phase) which minimizes delay induced by the controller. The feedback phase is comprised of everything that happens after the current state is measured. We sample the polytopes, linearize about the trajectory for every node but the first, and update the contact schedule in the preparation phase. In the feedback phase we compute the linearization for the first node, then we compute the MPC solution. We set the MPC period (i.e. the loop rate) at 10 ms, but the feedback phase generally takes much less time; see Table \ref{tab:mpc_timings} for a full break down. No line search was used.

Although the MPC is only computed at 100 Hz, a new feedforward torque and PD setpoint are output to the robot at 1 kHz by linearly interpolating the MPC output trajectory, in the same way as \cite{khazoom_tailoring_2024}. The output trajectory is interpolated at the current time, where the start time of the trajectory is set to when the state was measured. This means that if the MPC computation takes 5 ms, then the first control action from the new trajectory will be the action 5 ms into the trajectory. In this way we achieve a form of delay compensation \cite{bledt_implementing_2019}.

To increase robustness, we shrink the polytopes in the constraint to be smaller than the actual terrain. This helps guide the feet away from the edge. To prevent issues with infeasibility, we do not enforce the polytope constraint for any foot currently in contact. If the foot is in contact but not within the polytope, then the QP will be infeasible because the polytope constraint will directly contradict the no-slip constraint and, therefore, we choose not to enforce it. In a similar vein, the swing height constraint is not enforced until the third node to prevent any potential for infeasibility. This gives the foot a few nodes to go from its current height to the required height. 

As noted before, only a single SQP iteration is run, but we go further than that and fix the number of iterations in the QP solver. In this way, the QP does not finish solving but still provides us with usable results. For the quadruped we cap the iteration count within HPIPM to 10 and for the humanoid we cap it at 8. For both robots the ``speed" mode for HPIPM is used.

Unlike other methods \cite{grandia_perceptive_2022, khazoom_tailoring_2024, bledt_implementing_2019}, no adaptive time discretization between the nodes is used, although the MPC does use non-uniform discretization. The first 5 nodes are set to finer discretization than the rest. In quadruped experiments 25 nodes were used with the first 5 at a discretization of 0.015 s and the last 20 a discretization of 0.035 s. For the humanoid, the first 5 nodes had a 0.025 s discretization and the last 20 had 0.035 s discretization.

The cost function features a cost term guiding the end-effector positions. We use the Raibert heuristic with a moderate cost here to help guide the optimization solution. We found that the MPC works without this term too, but adding it appeared to help increase the robustness.

For the quadruped experiments we used the Unitree Go2 quadruped which has 12 actuators and 18 DoF. We assign one contact point to each foot for a total of 12 force variables per node. For the humanoid experiments we use the Unitree G1 humanoid which has 27 actuators (excluding the hands) and 33 DoF. We choose to only optimize over 19 actuators, fixing the others. Specifically, we fix both ankle roll joints and all the wrist joints. In simulation these joints are not fixed and a PD controller acts on them keeping them at a desired fixed position.

\section{Results}
To examine the benefits of the proposed controller, we will use simulation and hardware experiments. All simulation results are run in Mujoco asynchronously with the controller. The sensor measurements come into the estimator asynchronously and at the same rates as on hardware. All data is communicated through ROS2 to make the simulation as close to reality as possible. A PD controller is run in the Mujoco simulation to mimic the high-frequency motor controller running on the real robot. The simulation time step is 0.0005 seconds, and the implicitfast integrator with elliptic cones is used. Due to the extensive use of timing and the asynchronous nature of the simulation, the results are not deterministic.

\begin{figure}
    \vspace{3mm}    
    \centering
    \includegraphics[width=1\linewidth]{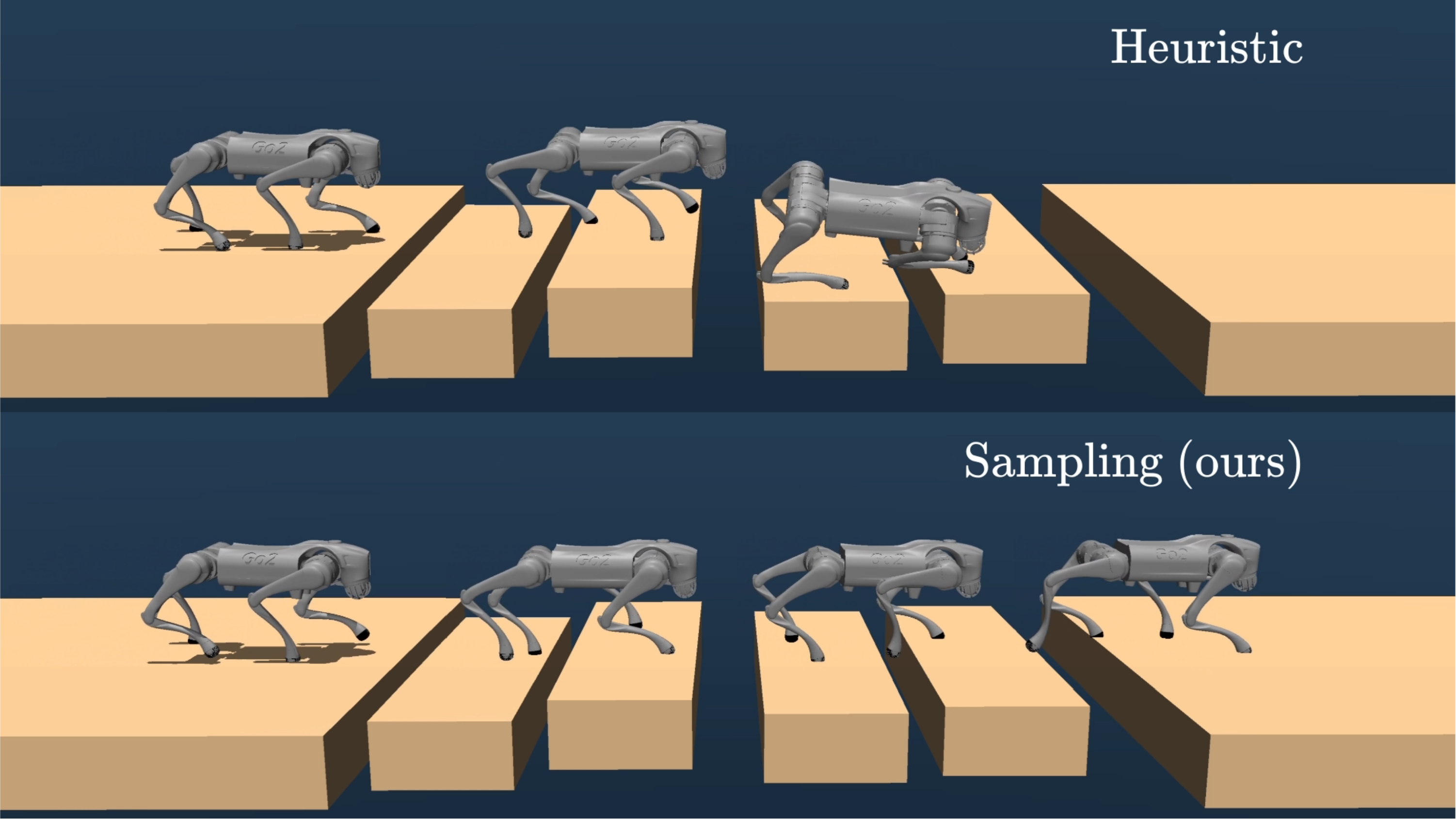}    \caption{Comparison between using our sampling based approach and a heuristic for terrain selection. The quadruped attempts to navigate over terrain with varying heights and gaps at 0.7 m/s. While using the heuristic, it is not able to stabilize towards the end of the path whereas the sampling method efficiently completes the entire path.}
    \label{fig:heuristic_comparison}
    \vspace{-3mm}    
\end{figure}

To compare the sampling method against the Raibert heuristic for terrain selection on the Go2 quadruped, we conduct a simulation experiment with stepping stones at varying heights. The quadruped is tasked with walking over the terrain at 0.7 m/s. For the sampling method, four samples are used and for the Raibert heuristic only a single MPC thread is used (as no sampling is required). Five simulations of the same experiment are run for each method to mitigate differences caused by the non-determinism of the simulation setup. We find that the sampling method outperforms the Raibert heuristic by a significant margin. In our experiments, the Raibert heuristic fails to reach the final platform 80\% of the time whereas the sampling method only fails once. For the sampling, the average cost of the successes was 0.8946 and for the heuristic it was 1.0963, showing an improvement in the optimality even across successes. An example run is shown in Fig. \ref{fig:heuristic_comparison}.

\begin{figure*}
    \vspace{2mm}
    \centering
    \includegraphics[width=1\linewidth]{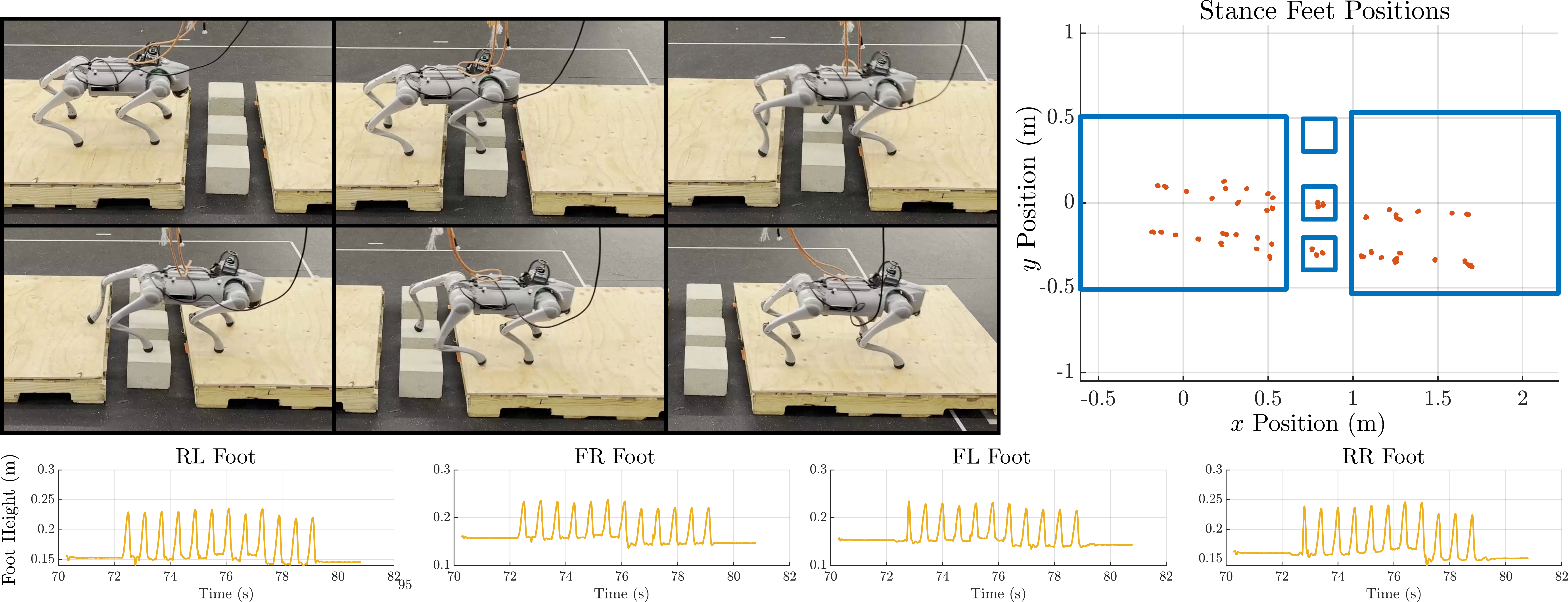}
    \caption{Hardware experiment on the Unitree Go2 quadruped where the terrain is composed of two large tiles separated by three smaller blocks. The quadruped navigates over the blocks smoothly, evaluating its options in real time using four samples. The top left shows snap shots from the hardware experiment. The top right plot shows where the feet were during the stance phase (in orange) and the polytopes representing the terrain (in blue). The bottom plots show the height of the feet throughout the experiment. The desired swing height was set to 8 cm above the measured stance height and the small blocks had a slightly different height than the large tiles.}
    \label{fig:quadruped_hardware}
\end{figure*}

Fig. \ref{fig:quadruped_hardware} shows a hardware experiment with the Unitree Go2 quadruped. The quadruped needs to traverse a gap where there are a few small cinder blocks to step on. Here we use four samples and see that the quadruped can successfully traverse the terrain. This demonstrates that the controller can run in the real world and is robust to model-mismatches. Beyond trivial model mismatch, we noticed that the large wooden platforms were not completely stationary and would wobble, thus providing other real-world disturbances. In addition, the lidar unit mounted on the quadruped was not modeled in the MPC (the lidar weight is about 265 g). Fig. \ref{fig:hero_fig} shows snapshots from a few other hardware experiments.

\begin{figure}
    \vspace{3mm}
    \centering
    \includegraphics[width=1\linewidth]{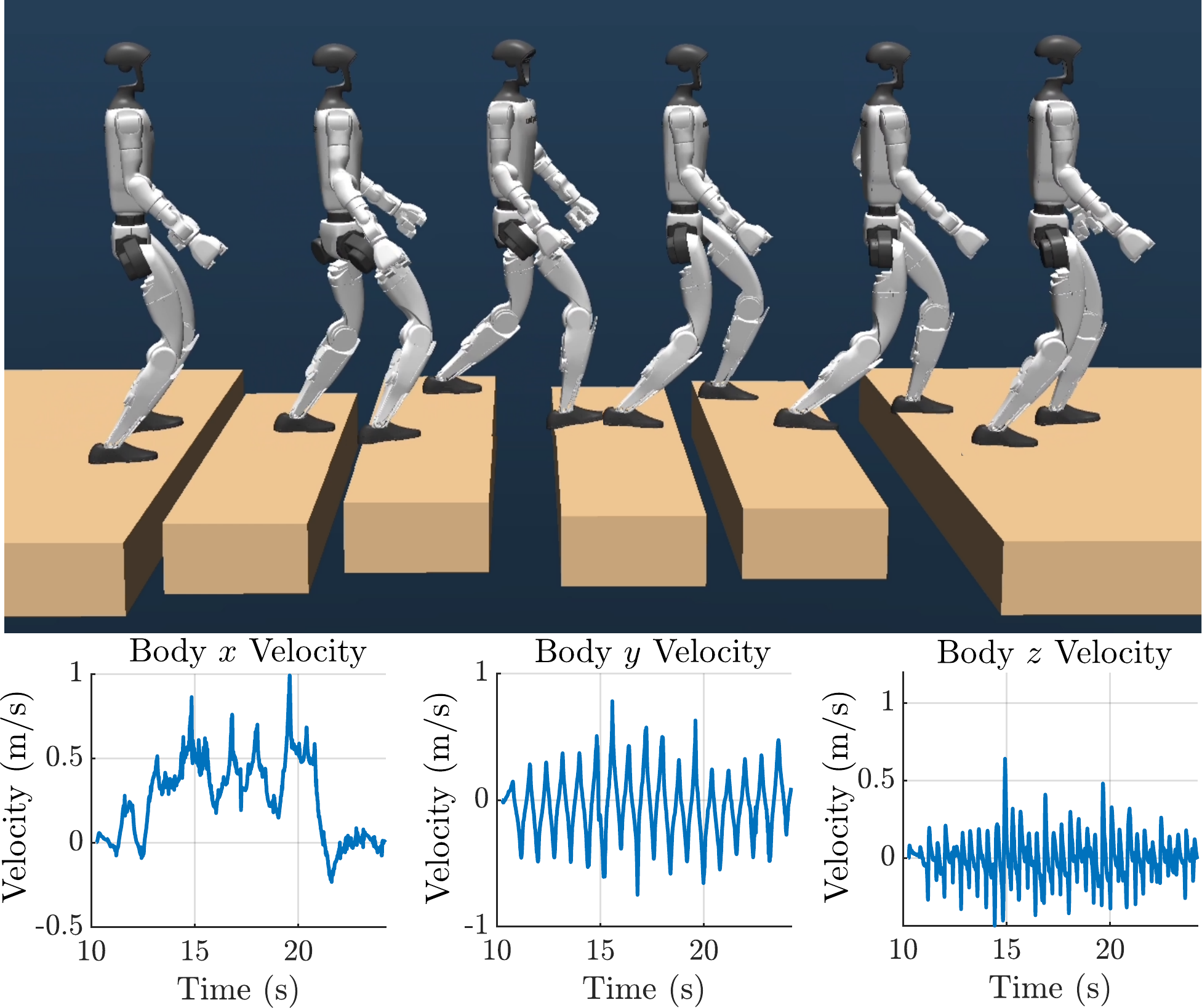}
    \caption{Simulation demonstration of the humanoid traversing stepping stones at varying heights. The humanoid operates with two samples to choose the terrain.}
    \label{fig:biped_walking}
    \vspace{-3mm}
\end{figure}

The proposed controller can also control a full humanoid robot in real time. A simulation of the Unitree G1 humanoid robot is run where the robot must traverse stepping stones with varying heights and gaps. Snapshots of the humanoid are shown in Fig. \ref{fig:biped_walking}. Here we use two samples and find that we can still run the controller in real time. The humanoid traverses the gaps naturally, and since the arms are being optimized, we can see that robot uses its arms and waist together to keep the torso straight even when the legs swing relatively fast to cross the gaps.

Table \ref{tab:mpc_timings} gives a full break down of the measured timings of the algorithm. These timings are gathered from tests where the robot walks over the stepping stones shown in Figs. \ref{fig:heuristic_comparison} and \ref{fig:biped_walking}. The algorithm was run on an Intel i9-14900K desktop CPU. Although direct timing comparisons are hard as they rely on the exact computer and engineering tricks used, we find that our algorithm is very performant, generally measuring faster than CIMPC approaches (see \cite{kurtz_inverse_2023} for a CIMPC timing comparison), and faster than full order sampling methods running on desktop GPUs \cite{kurtz_hydrax_2024, xue_full-order_2024}. The speed of the algorithm is shown to be fast enough for real-time control of both quadrupeds and humanoids through the hardware experiments and asynchronous simulations. 

\begin{table*}
    \vspace{3mm}
    \caption{MPC Timings}
    \begin{center}
    \begin{tabular}{| c | c | c | c | c | c |} 
     \hline
     Robot & Optimization DoF & Nodes & \# Parallel MPC & Prep Time (ms) & Feedback Time (ms) \\ 
     & & & & Average / Std. Dev. & Average / Std. Dev. \\
     \hline
        G1 & 25 & 25 & 1 & 1.730 / 0.120 & 5.525 / 0.383 \\
        G1 & 25 & 25 & 2 & 1.827 / 0.196 & 6.151 / 1.071 \\
        Go2 & 18 & 25 & 1 & 1.109 / 0.077 & 3.114 / 0.563 \\
        Go2 & 18 & 25 & 4 & 1.226 / 0.139 & 4.259 / 1.187 \\
     \hline
    \end{tabular}
    \end{center}
    \label{tab:mpc_timings}
    \vspace{-6mm}
\end{table*}

\section{Conclusion And Future Work}
We presented a layered controller with MPC for dynamic locomotion over constrained footholds. This controller demonstrates that through the use of both gradient-free and gradient-based methods, both discrete and continuous variables can be optimized in real time to improve performance. The controller preserves the speed of MPC while getting the optimality and generality of sampling based methods. We demonstrated these capabilities with hardware and simulation experiments for quadrupeds and bipeds.

The examples in this paper were specifically focused on the stepping stone problem, but we believe the presented architecture to be more broad than that. Future work should focus on using this for other tasks, such as using a hand to brace against a wall in push recovery, or optimizing the contact schedule. Further, there is room for experimentation with the sampling algorithm. We chose to sample directly at the terrain layer, but one could imagine that the sampling layer could be done with a full order optimizer like MPPI and the relevant contact information extracted out and passed to the gradient based MPC.

We used the idea of Real Time Iterations for the MPC computation, but we did not examine how the this paradigm interfaces with the changing constraints dictated by the sampling process. This becomes interesting when we don't solve the QP to completion as done here. Future work could examine this interaction and attempt to design a sampling method that is advantageous for this computing paradigm.  

\bibliographystyle{IEEEtran}
\bibliography{Layered_MPC_For_Locomotionv2}

\end{document}